\documentclass[11pt,letterpaper]{article}
\usepackage{naaclhlt2013}
\usepackage{latexsym}
\usepackage{times}
\usepackage{graphicx}
\usepackage{mathtools}
\usepackage{alltt}
\usepackage{url}
\usepackage{multirow}
\usepackage{float}
\usepackage{lipsum}
\usepackage{todonotes}
\usepackage{enumitem}
\usepackage{amsmath}
\restylefloat{table}

\providecommand{\e}[1]{\ensuremath{\times 10^{#1}}}
\title{ManTIME: Temporal expression identification and normalization in the TempEval-3 challenge}

\author{Michele Filannino, Gavin Brown, Goran Nenadic\\
	    The University of Manchester\\
	    School of Computer Science\\
	    Manchester, M13 9PL, UK\\
	    {\tt \{m.filannino, g.brown, g.nenadic\}@cs.man.ac.uk}
}

\date{}

\begin{document}
\maketitle
\begin{abstract}
This paper describes a temporal expression identification and normalization system, ManTIME, developed for the TempEval-3 challenge. The identification phase combines the use of conditional random fields along with a post-processing identification pipeline, whereas the normalization phase is carried out using NorMA, an open-source rule-based temporal normalizer.
We investigate the performance variation with respect to different feature types. Specifically, we show that the use of WordNet-based features in the identification task negatively affects the overall performance, and that there is no statistically significant difference in using gazetteers, shallow parsing and propositional noun phrases labels on top of the morphological features.
On the test data, the best run achieved 0.95 (P), 0.85 (R) and 0.90 (F1) in the identification phase. Normalization accuracies are 0.84 (type attribute) and 0.77 (value attribute). Surprisingly, the use of the silver data (alone or in addition to the gold annotated ones) does not improve the performance.
\end{abstract}

\section{Introduction}


Temporal information extraction~\cite{Verhagen07,Verhagen10} is pivotal for many Natural Language Processing (NLP) applications such as question answering, text summarization and machine translation. Recently the topic aroused increasing interest also in the medical domain~\cite{Sun12,Kovacevic13}.

Following the work of Ahn et al.~\shortcite{Ahn05}, the temporal expression extraction task is now conventionally divided into two main steps: identification and normalization. In the former step, the effort is concentrated on how to detect the right boundary of temporal expressions in the text. In the normalization step, the aim is to interpret and represent the temporal meaning of the expressions using TimeML~\cite{Pustejovsky03} format. In the TempEval-3 challenge~\cite{UzZaman12} the normalization task is focused only on two temporal attributes: \emph{type} and \emph{value}.

\section{System architecture}

ManTIME mainly consists of two components, one for the identification and one for the normalization.

%

\subsection{Identification}

We tackled the problem of identification as a sequencing labeling task leading to the choice of Linear Conditional Random Fields (CRF)~\cite{Lafferty01}. We trained the system using both human-annotated data (TimeBank and AQUAINT corpora) and silver data (TE3Silver corpus) provided by the organizers of the challenge in order to investigate the importance of the silver data.

Because the silver data are far more numerous (660K tokens vs. 95K), our main goal was to reinforce the human-annotated data, under the assumption that they are more informative with respect to the training phase. Similarly to the approach proposed by Adafre and de Rijke~\shortcite{Adafre05}, we developed a post-processing pipeline on top of the CRF sequence labeler to boost the results. Below we describe each component in detail.

\subsubsection{Conditional Random Fields}



The success of applying CRFs mainly depends on three factors: the labeling scheme (\emph{BI}, \emph{BIO}, \emph{BIOE} or \emph{BIOEU}), the topology of the factor graph and  the quality of the features used. We used the \emph{BIO} format in all the experiments performed during this research. The factor graph has been generated using the following topology: $(w_{0})$, $(w_{-1})$, $(w_{-2})$, $(w_{+1})$, $(w_{+2})$, $(w_{-2} \wedge w_{-1})$, $(w_{-1} \wedge w_{0})$, $(w_{0} \wedge w_{+1})$, $(w_{-1} \wedge w_{0} \wedge w_{+1})$, $(w_{0} \wedge w_{+1} \wedge w_{+2})$, $(w_{+1} \wedge w_{+2})$, $(w_{-2} \wedge w_{-1} \wedge w_{0})$, $(w_{-1} \wedge w_{+1})$ and $(w_{-2} \wedge w_{+2})$.

The system tokenizes each document in the corpus and extracts 94 features. These belong to the following four disjoint categories:

\begin{itemize}


\item {\bf Morphological}: This set includes a comprehensive list of features typical of Named Entity Recognition (NER) tasks, such as the word as it is, lemma, stem, pattern (e.g. \emph{'Jan-2003': 'Xxx-dddd'}), collapsed pattern (e.g. \emph{'Jan-2003': 'Xx-d'}), first 3 characters, last 3 characters, upper first character, presence of 's' as last character, word without letters, word without letters or numbers, and verb tense. For lemma and POS tags we use TreeTagger~\cite{TreeTagger}. Boolean values are included, indicating if the word is lower-case, alphabetic, digit, alphanumeric, titled, capitalized, acronym (capitalized with dots), number, decimal number, number with dots or stop-word. Additionally, there are features specifically crafted to handle temporal expressions in the form of regular expression matching: cardinal and ordinal numbers, times, dates, temporal periods (e.g. \emph{morning}, \emph{noon}, \emph{nightfall}), day of the week, seasons, past references (e.g. \emph{ago}, \emph{recent}, \emph{before}), present references (e.g. \emph{current}, \emph{now}), future references (e.g. \emph{tomorrow}, \emph{later}, \emph{ahead}), temporal signals (e.g. \emph{since}, \emph{during}), fuzzy quantifiers (e.g. \emph{about}, \emph{few}, \emph{some}), modifiers, temporal adverbs (e.g. \emph{daily}, \emph{earlier}), adjectives, conjunctions and prepositions.

\item {\bf Syntactic}: Chunks and propositional noun phrases belong to this category. Both are extracted using the shallow parsing software MBSP\footnote{http://www.clips.ua.ac.be/software/mbsp-for-python}.

\item {\bf Gazetteers}: These features are expressed using the BIO format because they can include expressions longer than one word. The integrated gazetteers are: male and female names, U.S. cities, nationalities, world festival names and ISO countries.

\item {\bf WordNet}: For each word we use the number of senses associated to the word, the first and the second sense name, the first 4 lemmas, the first 4 entailments for verbs, the first 4 antonyms, the first 4 hypernyms and the first 4 hyponyms. Each of them is defined as a separate feature.
\end{itemize}

The features mentioned above have been combined in 4 different models:
\begin{itemize}
\item {\bf Model 1}: Morphological only
\item {\bf Model 2}: Morphological + syntactic
\item {\bf Model 3}: Morphological + gazetteers
\item {\bf Model 4}: Morphological + gazetteers + WordNet
\end{itemize}

All the experiments have been carried out using CRF++ 0.57\footnote{https://code.google.com/p/crfpp/} with parameters \begin{math}C=1\end{math}, \begin{math}\eta=0.0001\end{math} and L2-regularization function.

\subsubsection{Model selection}

The model selection was performed over the entire training corpus. Silver data and human-annotated data were merged, shuffled at sentence-level (seed = 490) and split into two sets: 80\% as cross-validation set and 20\% as real-world test set. The cross-validation set was shuffled 5 times, and for each of these, the 10-fold cross validation technique was applied.

\begin{figure}[h]
\includegraphics[width=\linewidth]{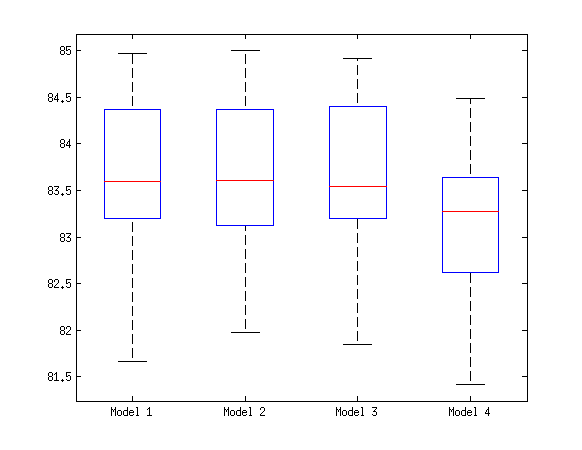}
\caption{Differences among models using 5x10-fold cross-validation}
\label{boxplot_crf}
\end{figure}
%

The analysis is statistically significant ($p = 0.0054$ with ANOVA test) and provides two important outcomes: (i) the set of WordNet features negatively affects the overall classification performance, as suggested by Rigo et al.~\shortcite{Rigo11}. We believe this is due to the sparseness of the labels: many tokens did not have any associated WordNet sense. (ii) There is no statistically significant difference among the first three models, despite the presence of apparently important information such as chunks, propositional noun phrases and gazetteers. The Figure \ref{boxplot_crf} shows the box plots for each model.

In virtue of this analysis, we opted for the smallest feature set (Model 1) to prevent overfitting.

In order to get a reliable estimation of the performance of the selected model on the real world data, we trained it on the entire cross-validation set and tested it against the real-word test set. The results for all the models are shown in the following table:
\begin{table}[h!]
\begin{center}
\begin{tabular}{|l|c|c|c|}
\hline \bf System & \bf Pre. & \bf Rec. & \bf F$_{\boldsymbol \beta=1}$ \\ \hline
Model 1 & 83.20 & \bf 85.22 & \bf 84.50 \\
Model 2 & \bf 83.57 & 85.12 & 84.33 \\
Model 3 & 83.51 & 85.12 & 84.31 \\
Model 4 & 83.15 & 84.44 & 83.79 \\
\hline
\end{tabular}
\end{center}
\end{table}

\begin{table*}
\begin{tabular}{|c|l||c|c|c|c|c|c||c|c||c|c|}
\hline
\multirow{2}{*}{\bf \#} & \multirow{2}{*}{\bf Training data} & \multicolumn{6}{|c||}{Identification} & \multicolumn{2}{|c||}{Normalization} & \multirow{2}{*}{\bf Overall} \\ \cline{3-10} 
& & \multicolumn{3}{|c|}{Strict matching} & \multicolumn{3}{|c||}{Lenient matching} & \multicolumn{2}{|c||}{Accuracy} & \\ \cline{3-10}
\bf run & \bf (post-processing) & \bf Pre. & \bf Rec. & $\bf F_{\boldsymbol\beta=1}$ & \bf Pre. & \bf Rec. & $\bf \tilde{F}_{\boldsymbol\beta=1}$ & \bf Type & \bf Value & {\bf score} \\ \hline
1 & Human\&Silver (no) & 78.57 & 63.77 & 70.40 & 97.32 & 78.99 & 87.20 & 88.99 & 77.06 & 67.20 \\
2 & Human\&Silver (yes) & 79.82 & 65.94 & 72.22 & 97.37 & 80.43 & 88.10 & 87.38 & 75.68 & 66.67 \\
3 & Human (no) & 76.07 & 64.49 & 69.80 & 94.87 & 80.43 & 87.06 & 87.39 & 77.48 & 67.45 \\
4 & Human (yes) & 78.86 & \bf 70.29 & \bf 74.33 & 95.12 & \bf 84.78 & \bf 89.66 & 86.31 & 76.92 & \bf 68.97 \\
5 & Silver (no) & 77.68 & 63.04 & 69.60 & 97.32 & 78.99 & 87.20 & 88.99 & 77.06 & 67.20 \\
6 & Silver (yes) & \bf 81.98 & 65.94 & 73.09 & \bf 98.20 & 78.99 & 87.55 & \bf 90.83 & \bf 77.98 & 68.27 \\
\hline
\end{tabular}
\caption{\label{performance} Performance on the TempEval-3 test set. }
\end{table*}

Precision, Recall and $F_{\beta=1}$ score are computed using strict matching.

The models used for the challenge have been trained using the entire training set. 

\subsubsection{Post-processing identification pipeline}
\label{postprocessing}

Although CRFs already provide reasonable performance, equally balanced in terms of precision and recall, we focused on boosting the baseline performance through a post-processing pipeline. For this purpose, we introduced 3 different modules.

{\bf Probabilistic correction module} averages the probabilities from the trained CRFs model with the ones extracted from human-annotated data only. For each token, we extracted: (i) the conditional probability for each label to be assigned (\emph{B}, \emph{I} or \emph{O}), and (ii) the prior probability of the labels in the human-annotated data only. The two probabilities are averaged for every label of each token. The list of tokens extracted in the human-annotated data was restricted to those that appeared within the span of temporal expressions at least twice. The application of this module in some cases has the effect of changing the most likely label leading to an improvement of recall, although its major advantage is making CRFs predictions less strict. 

{\bf BIO fixer} fixes wrong label sequences. For the BIO labeling scheme, the sequence \emph{O-I} is necessarily wrong. We identified \emph{B-I} as the appropriate substitution. This is the case in which the first token has been incorrectly annotated (e.g. \emph{``Three/O days/I ago/I ./O''} is converted into \emph{``Three/B days/I ago/I ./O''}). We also merged close expressions such as \emph{B-B} or \emph{I-B}, because different temporal expressions are generally divided at least by a symbol or a punctuation character (e.g. \emph{``Wednesday/B morning/B''} is converted into \emph{``Wednesday/B morning/I''}).

{\bf Threshold-based label switcher} uses the probabilities extracted from the human-annotated data. When the most likely label (in the human-annotated data) has a prior probability greater than a certain threshold, the module changes the CRFs predicted label to the most likely one. This leads to force the probabilities learned from the human-annotated data.

Through repeated empirical experiments on a small sub-set of the training data, we found an optimal threshold value (0.87) and an optimal sequence of pipeline components (Probabilistic correction module, BIO fixer, Threshold-based label switcher, BIO fixer).

We analyzed the effectiveness of the post-processing identification pipeline using a 10-fold cross-validation over the 4 models. The difference between CRFs and CRFs + post-processing pipeline is statistically significant ($p = 3.51\e{-23}$ with paired T-test) and the expected average increment is 2.27\% with respect to the strict $F_{\beta=1}$ scores.



\subsection{Normalization}
\label{sect:normalization}

The normalization component is an updated version of NorMA~\cite{Filannino12}, an open-source rule-based system.

\section{Results and Discussion}
\label{ssec:results}

We submitted six runs as combinations of different training sets and the use of the post-processing identification pipeline. The results are shown in Table~\ref{performance} where the \emph{overall score} is computed as multiplication between lenient $F_{\beta=1}$ score and the \emph{value} accuracy.


In all the runs, recall is lower than precision. This is an indication of a moderate lexical difference between training data and test data. The relatively low \emph{type} accuracy testifies the normalizer's inability to recognize new lexical patterns. Among the correctly typed temporal expressions, there is still about 10\% of them for which an incorrect \emph{value} is provided. The normalization task is proved to be challenging.


The training of the system by using human-annotated data only, in addition to the post-processing pipeline, provided the best results, although not the highest normalization accuracy. Surprisingly, the silver data do not improve the performance, both when used alone or in addition to human-annotated data (regardless of the post-processing pipeline usage).


The post-processing pipeline produces the highest precision when applied to the silver data only. In this case, the pipeline acts as a reinforcement of the human-annotated data. As expected, the post-processing pipeline boosts the performance of both precision and recall. We registered the best improvement with the human-annotated data.

Due to the small number of temporal expressions in the test set (138), further analysis is required to draw more general conclusions.


\section{Conclusions}
\label{sec:conclusions}

We described the overall architecture of ManTIME, a temporal expression extraction pipeline, in the context of TempEval-3 challenge.

This research shows, in the limits of its generality, the primary and exhaustive importance of morphological features to the detriment of syntactic features, as well as gazetteer and WordNet-related ones. In particular, while syntactic and gazetteer-related features do not affect the performance, WordNet-related features affect it negatively.

The research also proves the use of a post-processing identification pipeline to be promising for both precision and recall enhancement.

Finally, we found out that the silver data do not improve the performance, although we consider the test set too small for this result to be generalizable.

To aid replicability of this work, the system code, machine learning pre-trained models, statistical validation details and an online DEMO are available at: \url{http://www.cs.man.ac.uk/~filannim/projects/tempeval-3/}

%

\section*{Acknowledgments}

We would like to thank the organizers of the TempEval-3 challenge. The first author would like also to acknowledge Marilena Di Bari, Joseph Mellor and Daniel Jamieson for their support and the UK Engineering and Physical Science Research Council for its support in the form of a doctoral training grant.

\bibliography{tempeval-3_bibliography}

\begin{thebibliography}{}

\bibitem[\protect\citename{Adafre and de Rijke}2005]{Adafre05}
Sisay~Fissaha Adafre and Maarten de~Rijke.
\newblock 2005.
\newblock Feature engineering and post-processing for temporal expression
  recognition using conditional random fields.
\newblock In {\em Proceedings of the ACL Workshop on Feature Engineering for
  Machine Learning in Natural Language Processing}, FeatureEng '05, pages
  9--16, Stroudsburg, PA, USA. Association for Computational Linguistics.

\bibitem[\protect\citename{Ahn \bgroup et al.\egroup }2005]{Ahn05}
David Ahn, Sisay~Fissaha Adafre, and Maarten de~Rijke.
\newblock 2005.
\newblock Towards task-based temporal extraction and recognition.
\newblock In Graham Katz, James Pustejovsky, and Frank Schilder, editors, {\em
  Annotating, Extracting and Reasoning about Time and Events}, number 05151 in
  Dagstuhl Seminar Proceedings, Dagstuhl, Germany. Internationales Begegnungs-
  und Forschungszentrum f{\"u}r Informatik (IBFI), Schloss Dagstuhl, Germany.

\bibitem[\protect\citename{Filannino}2012]{Filannino12}
Michele Filannino.
\newblock 2012.
\newblock Temporal expression normalisation in natural language texts.
\newblock {\em CoRR}, abs/1206.2010.

\bibitem[\protect\citename{Kova{\'c}evi{\'c} \bgroup et al.\egroup
  }2013]{Kovacevic13}
Aleksandar Kova{\'c}evi{\'c}, Azad Dehghan, Michele Filannino, John~A Keane,
  and Goran Nenadic.
\newblock 2013.
\newblock Combining rules and machine learning for extraction of temporal
  expressions and events from clinical narratives.
\newblock {\em Journal of American Medical Informatics}.

\bibitem[\protect\citename{Lafferty \bgroup et al.\egroup }2001]{Lafferty01}
John~D. Lafferty, Andrew McCallum, and Fernando C.~N. Pereira.
\newblock 2001.
\newblock Conditional random fields: Probabilistic models for segmenting and
  labeling sequence data.
\newblock In {\em ICML}, pages 282--289.

\bibitem[\protect\citename{Pustejovsky \bgroup et al.\egroup
  }2003]{Pustejovsky03}
James Pustejovsky, Jos{\'e} Casta{\~n}o, Robert Ingria, Roser Saur{\'\i},
  Robert Gaizauskas, Andrea Setzer, and Graham Katz.
\newblock 2003.
\newblock Timeml: Robust specification of event and temporal expressions in
  text.
\newblock In {\em in Fifth International Workshop on Computational Semantics
  (IWCS-5}.

\bibitem[\protect\citename{Rigo and Lavelli}2011]{Rigo11}
Stefan Rigo and Alberto Lavelli.
\newblock 2011.
\newblock Multisex - a multi-language timex sequential extractor.
\newblock In {\em Temporal Representation and Reasoning (TIME), 2011 Eighteenth
  International Symposium on}, pages 163--170.

\bibitem[\protect\citename{Schmid}1994]{TreeTagger}
Helmut Schmid.
\newblock 1994.
\newblock Probabilistic part-of-speech tagging using decision trees.
\newblock In {\em Proceedings of the International Conference on New Methods in
  Language Processing}, Manchester, UK.

\bibitem[\protect\citename{Sun \bgroup et al.\egroup }2013]{Sun12}
Weiyi Sun, Anna Rumshisky, and Ozlem Uzuner.
\newblock 2013.
\newblock Evaluating temporal relations in clinical text: 2012 i2b2 challenge.
\newblock {\em Journal of the American Medical Informatics Association}.

\bibitem[\protect\citename{UzZaman \bgroup et al.\egroup }2012]{UzZaman12}
Naushad UzZaman, Hector Llorens, James~F. Allen, Leon Derczynski, Marc
  Verhagen, and James Pustejovsky.
\newblock 2012.
\newblock Tempeval-3: Evaluating events, time expressions, and temporal
  relations.
\newblock {\em CoRR}, abs/1206.5333.

\bibitem[\protect\citename{Verhagen \bgroup et al.\egroup }2007]{Verhagen07}
Marc Verhagen, Robert Gaizauskas, Frank Schilder, Mark Hepple, Graham Katz, and
  James Pustejovsky.
\newblock 2007.
\newblock Semeval-2007 task 15: Tempeval temporal relation identification.
\newblock In {\em Proceedings of the 4th International Workshop on Semantic
  Evaluations}, pages 75--80, Prague.

\bibitem[\protect\citename{Verhagen \bgroup et al.\egroup }2010]{Verhagen10}
Marc Verhagen, Roser Saur\'{\i}, Tommaso Caselli, and James Pustejovsky.
\newblock 2010.
\newblock Semeval-2010 task 13: Tempeval-2.
\newblock In {\em Proceedings of the 5th International Workshop on Semantic
  Evaluation}, SemEval '10, pages 57--62, Stroudsburg, PA, USA. Association for
  Computational Linguistics.

\end{thebibliography}

\end{document}